\documentclass[a4paper]{article}
\usepackage[dvips]{graphicx}
\usepackage{epsf}
\usepackage[fleqn]{amsmath}

\usepackage{amssymb}
\usepackage{amsbsy}
\usepackage{amsthm}

\newcommand{\norm}[1]{\lVert#1\rVert}

\newcommand{\jbf}[1]{{\boldsymbol{#1}}}

\newcommand{\Real}{{\mathbb R}}

\newcommand{\inner}[2]{\langle#1,#2\rangle}

\newcommand{\tphi}{\varphi}

\DeclareMathOperator{\argmin}{arg\,min}

\newcommand{\handout}[4]{
\renewcommand{\thepage}{\footnotesize Semi-Supervised Dimensionality Reduction
for Manifold Learning - \arabic{page}}

\begin{center}
{\Large #3}\\
\bigskip
\noindent {#1}

\noindent{#4}

\bigskip
\noindent {#2}

\end{center}

}

\begin{document}
\handout{Ratthachat Chatpatanasiri and Boonserm Kijsirikul}{May 10,
2009}{\textbf{A Unified Semi-Supervised Dimensionality Reduction Framework for Manifold Learning}}{Department of Computer Engineering, Chulalongkorn University,\\
Bangkok 10330, THAILAND.\\
\textsf{ratthachat.c@gmail.com} and \textsf{boonserm.k@chula.ac.th}}

\begin{center}
\subsection*{Abstract}
\end{center}
We present a general framework of semi-supervised dimensionality
reduction for manifold learning which naturally generalizes existing
supervised and unsupervised learning frameworks which apply the
spectral decomposition. Algorithms derived under our framework are
able to employ both labeled and unlabeled examples and are able to
handle complex problems where data form separate clusters of
manifolds. Our framework offers simple views, explains relationships
among existing frameworks and provides further extensions which can
improve existing algorithms. Furthermore, a new semi-supervised
kernelization framework called ``KPCA trick'' is proposed to handle
non-linear problems.\\

\noindent \textbf{Keywords:} Semi-supervised Learning, Transductive
Learning, Spectral Methods, Dimensionality Reduction, Manifold Learning, KPCA Trick.\\

\section{Introduction}
%Dimensionality reduction is one of the most important research areas
%of machine learning.

In many real-world applications, high-dimensional data indeed lie on
(or near) a low-dimensional subspace. The goal of
\emph{dimensionality reduction} is to reduce complexity of input
data while some desired intrinsic information of the data is
preserved. The desired information can be discriminative
\cite{Yan:PAMI07,Wei:ICML07,Cai:CVPR07,Hoi:CVPR06,Chen:CVPR05,Cheng:ICIP04},
geometrical
\cite{Tenenbaum:Science00,Roweis:Science00,He:NIPS04,Saul:SSL06} or
both \cite{Sugiyama:JMLR07}.
%There are several advantages of
%reducing the dimensionality of input data. First, working on a
%low-dimensional space significantly saves both time and storage.
%Second, an intuitive visualization is possible for low-dimensional
%data. Finally, and most importantly, working on a low-dimensional
%subspace secures us from the \emph{curse of dimensionality}
%\cite{Bishop:Book06}; ``learning'' is possible even if we have only
%a relatively few number of input data.
Fisher discriminant analysis (FDA) \cite{Fukunaga:Book90} is the
most popular method among all supervised dimensionality reduction
algorithms. Denote $c$ as the number of classes in a given training
set. Provided that training examples of each class lie in a linear
subspace and \emph{do not} form several separate clusters, i.e.
\emph{do not} form multi-modality, FDA is able to discover a
low-dimensional linear subspace (with at most $c-1$ dimensionality)
which is efficient for classification. Recently, many works have
improved the FDA algorithm in several aspects
\cite{Sugiyama:JMLR07,Yan:PAMI07,Wei:ICML07,Cai:CVPR07,Hoi:CVPR06,Chen:CVPR05,Cheng:ICIP04}.
These \emph{extended FDA algorithms} are able to discover a nice
low-dimensional subspace even when training examples of each class
lie in separate clusters of complicated non-linear manifolds.
Moreover, a subspace discovered by these algorithms has no
limitation of $c-1$ dimensionality.

Although the extended FDA algorithms work reasonably well, a
considerable number of labeled examples is required to achieve
satisfiable performance. In many real-world applications such as
image classification, web page classification  and protein function
prediction, a labeling process is costly and time consuming; in
contrast, unlabeled examples can be easily obtained. Therefore, in
such situations, it can be beneficial to incorporate the information
which is contained in unlabeled examples into a learning problem,
i.e., semi-supervised learning (SSL) should be applied instead of
supervised learning \cite{Semi:BOOK06}.

In this paper, we present a general semi-supervised dimensionality
reduction framework which is able to employ information from both
labeled and unlabeled examples. Contributions of the paper can be
summarized as follows.

$\bullet \ $ As the extended FDA algorithms, algorithms developed in
our framework are able to discover a nice low-dimensional subspace
even when training examples of each class form separate clusters of
complicated non-linear manifolds. In fact, those previous supervised
algorithms can be casted as instances in our framework. Moreover,
our framework explains previously unclear relationships among
existing algorithms in a simple viewpoint.

$\bullet \ $ We present a novel technique called the \emph{Hadamard
power operator} which improves the use of unlabeled examples in
previous algorithms. Experiments show that the Hadamard power
operator improves the classification performance of a
semi-supervised learner derived from our framework.

$\bullet \ $ We show that  recent existing semi-supervised
frameworks applying spectral decompositions known to us
\cite{Sugiyama:pakdd08,Song:PR08} can be viewed as special cases of
our framework. Moreover, empirical evidence shows that
semi-supervised learners derived from our framework are superior to
existing learners in many standard problems.

$\bullet \ $ A new non-linearization framework, namely, a \emph{KPCA
trick} framework \cite{Me:Arxiv08a} is extended into a
semi-supervised learning setting. In contrast to the standard kernel
trick, the KPCA trick does not require users to derive new
mathematical formulas and to re-implement the kernel version of the
original learner.

%The KPCA-trick framework also avoids problems such as singularity in
%eigen-decomposition, etc. Representer theorems are presented to
%prove the validation of the KPCA trick in the context of
%semi-supervised learning.
%
%$\bullet \ $ We conduct extensive experiments in order to show that
%semi-supervised learners derived from our framework have higher
%predictive abilities than those of existing learners in standard
%datasets.

\section{The Framework} \label{sect_framework}
Let $\{\textbf{x}_i, y_i\}_{i=1}^\ell$ denote a training set of
$\ell$ labeled examples, with inputs $\textbf{x}_i \in
{\mathbb{R}}^{d_0}$ generated from a fixed but unknown probability
distribution  $\mathcal{P}_\textbf{x}$, and corresponding class
labels $y_i \in \{1, ..., c\}$ generated from
$\mathcal{P}_{y|\textbf{x}}$. In addition to the labeled examples,
let $\{\textbf{x}_i\}_{i=\ell+1}^{\ell+u}$ denote a set of $u$
unlabeled examples also generated from $\mathcal{P}_\textbf{x}$.
Denote $X \in \Real^{{d_0} \times (\ell+u)}$ as a matrix of the
input examples $(\textbf{x}_1, ..., \textbf{x}_{\ell+u})$. We define
$n = \ell + u$. The goal of semi-supervised learning (SSL) dimensionality reduction is\\

\noindent \textbf{Goal.} Using the information of both labeled and
unlabeled examples, we want to map $(\textbf{x} \in
{\mathbb{R}}^{d_0}) \mapsto (\textbf{z} \in {\mathbb{R}}^{d})$ where
$d < d_0$, such that in the embedded space
$\mathcal{P}_{y|\textbf{z}}$
can be accurately estimated ( i.e., unknown labels are easy to predict) by \emph{a simple classifier}.\\

\noindent Here, following the previous works in the supervised
setting \cite{Sugiyama:JMLR07,Yan:PAMI07,Wei:ICML07}, the
\emph{nearest neighbor} algorithm is used for representing a simple
classifier mentioned in the goal. Note that important special cases
of SSL problems are \emph{transductive} problems where we only want
to predict the labels $\{y_i\}_{i=\ell+1}^{\ell+u}$ of the given
unlabeled examples. In order to make use of unlabeled examples in
the learning process, we make the following so-called \emph{manifold
assumption} \cite{Semi:BOOK06}:\\

\noindent \textbf{Semi-Supervised Manifold Assumption}. The support
of $\mathcal{P}_\mathbf{x}$ is on a low-dimensional manifold.
Furthermore, $\mathcal{P}_{y|\mathbf{x}}$ is smooth, as a function
of $\mathbf{x}$, with respect to the underlying structure of the
manifold.\\

%\noindent  In words,  this assumption states that two nearby points
%on a high-density region of $\mathcal{P}_\mathbf{x}$ are likely to
%be in the same class. Since unlabeled examples can be used to
%estimate $\mathcal{P}_\mathbf{x}$, they are also useful to predict
%the label of an example\footnote{ In fact, beside the manifold
%assumption, here we make an additional assumption that available
%unlabeled examples are generated from a high-density region of
%$\mathcal{P}_\mathbf{x}$. Nevertheless, this additional assumption
%is not too strong as it is implied by the law of large number.}.

At first, to fulfill our goal, we linearly parameterize
$\textbf{z}_i = A \textbf{x}_i$ where $A \in \Real^{d \times
{d_0}}$. Thus, $AX = (A\textbf{x}_1, ..., A\textbf{x}_n) \in
\Real^{d \times n}$ is a matrix of embedded points. An efficient
non-linear extension is presented in Section~\ref{sect_kpca}. In our
framework, we propose to cast the problem as a constrained
optimization problem:
\begin{equation} \label{eq_framework1}
  A^* = \underset{A \in \mathcal{A}}{\operatorname{\argmin}}\,
  f^{\ell}(AX) + \gamma f^{u}(AX),
\end{equation}
where $f^{\ell}(\cdot)$ and $f^{u}(\cdot)$ are objective functions
based on labeled and unlabeled examples, respectively, $\gamma$ is a
parameter controlling the weights between the two objective
functions and $\mathcal{A}$ is a constraint set in $\Real^{d \times
d_0}$. The two objective functions determine ``how good the embedded
points are''; thus, their arguments are $AX$, a matrix of embedded
points. Up to \emph{orthogonal and translational transformations},
we can identify embedded points via their pairwise distances instead
of their individual locations. Therefore, we can base the objective
functions on pairwise distances of embedded examples. Here, we
define the objective functions to be \emph{linear} with respect to
the pairwise distances:
\begin{align*}
  f^{\ell}(AX) = \sum_{i,j=1}^n c^\ell_{ij} \mbox{ dist}(A\textbf{x}_i,A\textbf{x}_j) \mbox{ and }
  f^{u}(AX) = \sum_{i,j=1}^n c^u_{ij} \mbox{ dist}(A\textbf{x}_i,A\textbf{x}_j),
\end{align*}
where $\mbox{dist}(\cdot,\cdot)$ is an arbitrary distance function
between two embedded points, $c^\ell_{ij}$ and $c^u_{ij}$ are costs
which penalize an embedded distance between two points $i$ and $j$.
A specification of $c^\ell_{ij}$ and $c^u_{ij}$ are based on the
label information and unlabel information, respectively, as
described in Section~\ref{sect_cost}.

\begin{figure}[h]
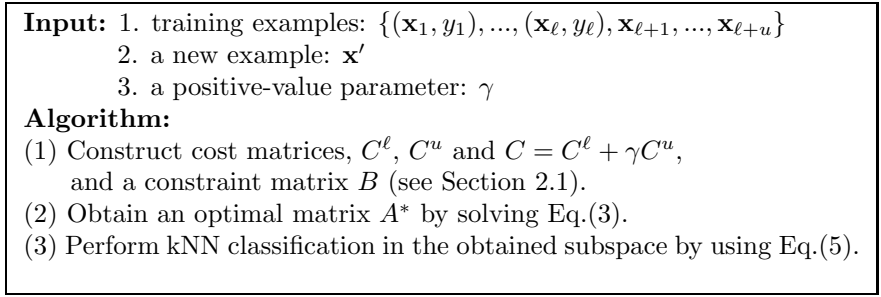

\begin{center}
\vskip -0.2in \fbox{
\begin{minipage}{11cm}
  \textbf{Input:} 1. training examples: $\{(\textbf{x}_1,y_1), ..., (\textbf{x}_\ell,y_\ell),\textbf{x}_{\ell+1}, ..., \textbf{x}_{\ell+u}\}$\\
  \hspace*{1.12cm} 2. a new example: $\textbf{x}'$\\
  \hspace*{1.12cm} 3. a positive-value parameter: $\gamma$\\
\textbf{Algorithm:}\\
    (1) Construct cost matrices, $C^\ell$, $C^u$ and $C = C^\ell + \gamma C^u$,\\
    \hspace*{0.5cm} and a constraint matrix $B$ (see Section~\ref{sect_cost}).\\
    (2) Obtain an optimal matrix $A^*$ by solving Eq.\eqref{eq_framework3}.\\
    (3) Perform kNN classification in the obtained subspace by using Eq.\eqref{eq_linear4}.\\
\end{minipage}
} \vskip -0.1in \caption{Our semi-supervised learning framework.}
\label{fig_ssl}
\end{center}
\vskip -0.2in
\end{figure}

If we restrict ourselves to consider only the cases that (I)
$\mbox{dist}(\cdot,\cdot)$ is a squared Euclidean distance function,
i.e. $\mbox{dist}(A\textbf{x}_i,A\textbf{x}_j) = \norm{A\textbf{x}_i
-A\textbf{x}_j}^2$, (II) $c^\ell_{ij}$ and $c^u_{ij}$ are symmetric,
and (III) $A \in \mathcal{A}$ is in the form of $A B A^T = I$ where
$B$ is a positive semidefinite (PSD) matrix,
Eq.\eqref{eq_framework1} will result in a general framework which
indeed generalizes previous frameworks as shown in
Section~\ref{sect_related}. Define $c_{ij} = c^\ell_{ij} + \gamma
c^u_{ij}$. We can rewrite the weighted combination of the objective
funtions in Eq.\eqref{eq_framework1} as follows:
\begin{align*}
 f^{\ell}&(AX) + \gamma f^{u}(AX) \\
 &= \sum_{i,j=1}^n c^\ell_{ij} \mbox{dist}(A\textbf{x}_i,A\textbf{x}_j) + \gamma \sum_{i,j=1}^n c^u_{ij}
\mbox{ dist}(A\textbf{x}_i,A\textbf{x}_j)\\
 &=\sum_{i,j=1}^n (c^\ell_{ij} + \gamma c^u_{ij})
 \mbox{dist}(A\textbf{x}_i,A\textbf{x}_j) =\sum_{i,j=1}^n c_{ij} \mbox{dist}(A\textbf{x}_i,A\textbf{x}_j)\\
 &= \sum_{i,j=1}^n c_{ij} \norm{A\textbf{x}_i - A\textbf{x}_j}^2
  = \ 2  \sum_{i,j=1}^n c_{ij} (\textbf{x}_i^T A^T A \textbf{x}_i - \textbf{x}_i^T A^T A \textbf{x}_j)\\
    &= \ 2 \mbox{trace} \left(A \Big(\sum_{i,j=1}^n (\textbf{x}_i c_{ij} \textbf{x}_i^T) - \sum_{i,j=1}^n(\textbf{x}_i
    c_{ij}\textbf{x}_j^T)\Big) A^T \right)\\
    &= \ 2\mbox{trace}(AX(D-C)X^TA^T),
\end{align*}
where $C$ is a symmetric cost matrix with elements $c_{ij}$ and $D$
is a diagonal matrix with $D_{ii} = \sum_{j} c_{ij}$\footnote{To
simplify our notations, in this paper whenever we define a cost
matrix $C'$ having elements $c'_{ij}$, we always define its
associated diagonal matrix $D'$ with elements $D'_{ii} = \sum_{j}
c'_{ij}$.}. Thus, the optimization problem (\ref{eq_framework1}) can
be restated as
\begin{equation} \label{eq_linear2}
  A^* = \underset{A B A^T = I}{\operatorname{\argmin}}\, \mbox{trace}(A X (D-C) X^T
  A^T),
\end{equation}
Note that the constraint $A B A^T = I$ prevents trivial solutions
such as every $A\textbf{x}_i$ is a zero vector. If $B$ is a positive
definite (PD) matrix, a solution of the above problem is given by
the bottom $d$ eigenvectors of the following generalized eigenvalue
problem \cite{Fukunaga:Book90,{vonLuxburg:STATCOMP07}}
\begin{equation} \label{eq_framework3}
  X(D-C)X^T \textbf{a}^{(j)} = \lambda_j B \textbf{a}^{(j)}, \quad j = 1, ...,d.
\end{equation}
Then the optimal linear map is
\begin{equation} \label{eq_framework4}
A^* = (\textbf{a}^{(1)}, ..., \textbf{a}^{(d)})^T.
\end{equation}
Note that, in terms of solutions of Eq.\eqref{eq_framework3}, it is
more convenient to represent $A^*$ by its rows $\textbf{a}^{(i)}$
than its columns $\textbf{a}_i$. Moreover, note that
\begin{equation} \label{eq_linear4}
  \norm{\textbf{z}-\textbf{z}'} = \norm{A^*\textbf{x}~-~A^*\textbf{x}'}.
\end{equation}
Therefore, kNN in the embedded space can be performed. Consequently,
an algorithm implemented under our framework consists of three steps
as shown in Figure~\ref{fig_ssl}.\\
% plus one optional step
%for non-linearization using the KPCA trick described in the next section:\\

\subsection{Specification of the Cost and Constraint Matrices} \label{sect_cost}
In this section, we present various reasonable approaches of
specifying the two cost matrices, $C^\ell$ and $C^u$, and the
constraint matrix, $B$, by using the label and unlabel information.
We use the two words ``unlabel information'' and ``neighborhood
information'' interchangeably in this paper.
%The label information has been used in every supervised learning
%algorithm where the neighborhood information has been used for a
%manifold learning algorithm.

\subsubsection{The Cost Matrix $C^\ell$ and the Constraint Matrix $B$}
Normally, based on the label information, classical supervised
algorithms usually require an embedded space to have the following
two desirable
conditions:\\

\noindent (1) two examples of the same
class stay close to one another, and\\
\noindent (2) two examples of different
classes stay far apart.\\

\noindent The two conditions are imposed in classical works such as
FDA. However, the first condition is too restrictive to capture
manifold and multi-modal structures of data which naturally arise in
some applications. Thus, the first condition should be relaxed as follows:\\

\begin{figure*}[t]
\begin{center}
\vskip -0.5in \setlength{\epsfxsize}{2.5in}
\centerline{\epsfbox{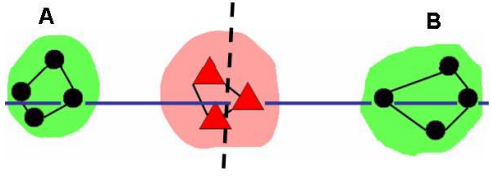}} \caption{An example when data
form a multi-modal structure. An algorithm, e.g. FDA, which imposes
the condition (1) will try to discover a new subspace (a dashed
line) which merges two clusters \textbf{A} and \textbf{B}
altogether. An obtained space is undesirable as data of the two
classes are mixed together. In contrast, an algorithm which imposes
the condition (1*) (instead of (1)) will discover a subspace (a
thick line) which does not merge the two clusters \textbf{A} and
\textbf{B} as there are no nearby examples (indicated by a link
between a pair of examples) between the two clusters.}
\label{fig_nearby} \vskip -0.3in
\end{center}
\end{figure*}

\noindent (1*) two \emph{nearby
examples} of the same class stay close to one another\\

\noindent where ``nearby examples'', defined by using the
neighborhood information, are examples which should stay close to
each other in both original and embedded spaces. The specification
of ``nearby examples'' has been proven to be successful in
discovering manifold and multi-modal structure
\cite{Sugiyama:JMLR07,Yan:PAMI07,Wei:ICML07,Cai:CVPR07,Hoi:CVPR06,Chen:CVPR05,Cheng:ICIP04,Goldberger:NIPS05,Globerson:NIPS06,Weinberger:NIPS06,Yang:AAAI06,Torresani:NIPS07}.
See Figure~\ref{fig_nearby} for explanations. In some cases, it is
also appropriate to relax the second condition to\\

\noindent (2*) two \emph{nearby examples} of different
classes stay far apart.\\

%\resizebox{25mm}{!}
% {\includegraphics{isomap.eps}}

In this section, we give three examples of cost matrices which
satisfy the conditions (1*) and (2) (or (2*)). These three examples
are recently introduced in previous works, namely, Discriminant
Neighborhood Embedding (DNE) \cite{Wei:ICML07}, Marginal Fisher
Analysis (MFA) \cite{Yan:PAMI07} and Local Fisher Discriminant
Analysis (LFDA) \cite{Sugiyama:JMLR07}, with different presentations
and motivations but they can be unified under our general framework.

Firstly, to specify nearby examples, we construct two matrices $C^I$
and $C^E$ based on any sensible distance (Euclidean distance is the
simplest choice). For each $\textbf{x}_i$, let $Neig^I(i)$ be the
set of $k$ nearest neighbors having the \emph{same} label $y_i$, and
let $Neig^E(i)$ be the set of $k$ nearest neighbors having
\emph{different} labels from $y_i$. Define $C^I$ and $C^E$ as
follows: let $c_{ij}^I = c_{ij}^E = 0$ if points $\textbf{x}_i$
and/or $\textbf{x}_j$ are unlabeled, and
\begin{equation*}
c^I_{ij} =
\begin{cases}
   1, & \mbox{if } j \in Neig^I(i) \vee i \in Neig^I(j), \\
   0, & \mbox{otherwise, and}
\end{cases}
\end{equation*}
\begin{equation*}
c^E_{ij} =
\begin{cases}
   1, & \mbox{if } j \in Neig^E(i) \vee i \in Neig^E(j), \\
   0, & \mbox{otherwise}.
\end{cases}
\end{equation*}
The specification $c^I_{ij} = 1$ and $c^E_{ij} = 1$ represent nearby
examples in the conditions (1*) and (2*). Then, $C^\ell$ and $B$ of
existing algorithms (see Eq.~\eqref{eq_linear2}) are:\\\\
\noindent \textbf{Discriminant Neighborhood Embedding (DNE)}\\
$C^\ell = C^I - C^E \quad \quad B = I$ (an identity matrix)\\

\noindent\textbf{Marginal Fisher Analysis (MFA)}\\
$C^\ell = -C^E \quad \quad \quad \quad B = X(D^I - C^I)X^T$\\

\noindent\textbf{Local Fisher Discriminant Analysis (LFDA)}\\
Let $n_1, ..., n_c$ be the numbers of examples of classes $1, ...,
c$, respectively. Define matrices $C^{bet}$ and $C^{wit}$ as:
\begin{equation*} \mathindent=0mm
\hspace*{0mm}
c^{bet}_{ij} =
\begin{cases}
   c^I_{ij}(\frac{1}{n_k} - \frac{1}{n}), & \mbox{if } y_i = y_j = k,\\
   -\frac{1}{n}, & \mbox{otherwise},
\end{cases} \ \ \mbox{and} \ \
c^{wit}_{ij} =
\begin{cases}
   \frac{1}{n_k}c^I_{ij}, & \mbox{if } y_i = y_j = k, \\
   0, & \mbox{otherwise},
\end{cases}
\end{equation*} \mathindent=7mm
%\begin{equation*}
%\end{equation*}
$C^\ell = C^{bet} \quad \quad \quad \quad \ \ B = X (D^{wit}-C^{wit}) X^T$\\

Within our framework, relationships among the three previous works
can be explained. The three methods exploit different ideas in
specifying matrices $C^\ell$ and $B$ to satisfy two desirable
conditions in an embedded space. In DNE, $C^\ell$ is designed to
penalize an embedded space which does not satisfy the condition (1*)
and (2*). In MFA, the constraint matrix $B$ is designed to satisfy
the condition (1*) and $C^\ell$ is designed to penalize an embedded
space which does not satisfy the condition (2*).
%\footnote{Note that we reformulate the
%original form of MFA presented in \cite{Yan:PAMI07} so that the
%problem can be conveniently handled when $B$ is PSD but not PD (see
%Section~\ref{sect_psd}). In its original form, the constraint matrix
%$B$ is designed to satisfy the condition (2*) and $C^\ell$ is
%designed to penalize a space which does not satisfy the condition
%(1*).}.

Things are not quite obvious in the case of LFDA. In LFDA, the
constraint matrix $B$ is designed to satisfy the condition (1*)
since elements $C^{wit}$ are proportional to $C^{I}$; nevertheless,
since weights are inversely proportional to $n_k$, elements in a
small class have larger weights than elements in a bigger class,
i.e. a pair in a small class is more likely to satisfy the condition
(1*) than a pair in a bigger class. To understand $C^\ell$, we
recall that\mathindent=0mm
\begin{align*}
\mbox{trace}(AX(D^{\ell} -  C^{\ell})X^TA^T)
  &= \sum_{i,j} c^\ell_{ij}\norm{A\textbf{x}_i - A\textbf{x}_j}\\
  &= \sum_{y_i = y_j} c^I_{ij}(\frac{1}{n_k} - \frac{1}{n})\norm{A\textbf{x}_i -
        A\textbf{x}_j} -
     \sum_{y_i \neq y_j} \frac{1}{n}\norm{A\textbf{x}_i -
        A\textbf{x}_j}\\
  &= d -\frac{1}{n} \left( \sum_{y_i = y_j} c^I_{ij}\norm{A\textbf{x}_i -
        A\textbf{x}_j} +
     \sum_{y_i \neq y_j} \norm{A\textbf{x}_i -
        A\textbf{x}_j}\right),
\end{align*} \mathindent=12mm
where at the third equality we use the constraint $AX B X^TA^T = I$
and hence
\begin{align*}
\mbox{trace}(AX(D^{wit} -  C^{wit})X^TA^T)
  &= \sum_{y_i = y_j} \frac{c^I_{ij}}{n_k}
        \norm{A\textbf{x}_i - A\textbf{x}_j}\\
  &= \mbox{trace}(I) = d.
\end{align*}
Hence, we observe that \emph{every} pair of labeled examples coming
from \emph{different} classes has a corresponding cost of
$-\frac{1}{n}$. Therefore, $C^\ell$ is designed to penalize an
embedded space which does not satisfy the condition (2).
Surprisingly, in LFDA, nearby examples of the \emph{same} class
(having $c^I_{ij} = 1$) also has a cost of $-\frac{1}{n}$. As a cost
proportional to $-\frac{1}{n}$ is meant to preserve a pairwise
distance between each pair of examples (see
Section~\ref{sect_self}), LFDA tries to preserve a local geometrical
structure between each pair of nearby examples of the same class, in
contrast to DNE and MFA which try to squeeze nearby examples of the
same class into a single point

We note that other recent supervised methods for manifold learning
can also be presented and interpreted in our framework with
different specifications of $C^\ell$, for examples, \emph{Local
Discriminant Embedding} of Chen et al. \cite{Chen:CVPR05} and
\emph{Supervised Nonlinear Local Embedding} of Cheng et al
\cite{Cheng:ICIP04}.

\subsubsection{The Cost Matrix $C^u$ and the Hadamard Power Operator} \label{sect_unlabel}
One important implication of the manifold assumption is that
``nearby examples are likely to belong to a same class''. Hence, by
the assumption, it makes sense to design $C^u$ such that \emph{it
prevents any pairs of nearby examples to stay far apart in an
embedded space}.

%As $C^I$ and $C^E$ defined in the previous subsection, the 0/1, i.e.
%``hard'', specification of nearby examples can be employed. However,
%the 0/1 specification can be thought of as discretization of the
%neighborhood information and thus some information can be loss (only
%1 bit is needed to store `0' or `1'), e.g. 0.5372 may be truncated
%to either 0 or 1 depending on a truncation scheme. In contrast to
%$C^\ell$ where we have the label information which is central in classification tasks, we do not have any
%additional information to specify $C^u$. Therefore, it would be
%appropriate to prevent the loss of information by presenting the
%neighborhood information in real value, i.e. a ``soft''
%specification. In this section, various possible choices of $C^u$
%which are employed in previous works of unsupervised manifold
%learning are presented under our framework.

%This subsection surveys various possible choices for constructing
%$W^{(2)}$. First, remember that we construct $W^{(2)}$ for
%approximating $\mathcal{P}_\textbf{x}$. Since labeled examples are
%also assumed to be generated from $\mathcal{P}_\textbf{x}$, we can
%also use labeled examples, together with unlabeled ones, to
%construct $W^{(2)}$.

Among methods of extracting the neighborhood information to define
$C^u$, methods based on \emph{the heat kernel} (or \emph{the
gaussian function}) are most popular. Beside using the heat kernel,
other methods of defining $C^u$ are invented, see \cite[Chap.
15]{Semi:BOOK06} and \cite{vonLuxburg:STATCOMP07} for more details.
The simplest specifications of nearby examples based on the heat
kernel are:
\begin{equation} \label{eq_globalscale}
c^{u}_{ij} =
\exp(\frac{\norm{\textbf{x}_i-\textbf{x}_j}^2}{\sigma^2} ).
\end{equation}
Each pair of nearby examples will be penalized with different costs
depended on their similarity, and a similarity between two points is
based on the Euclidean distance between them in the input space.
Incidentally, with this specification of $C^u$, the term $f^u(AX)$
in Eq.~\eqref{eq_framework1} can be interpreted as an approximation
of the \emph{Laplace-Beltrami operator} on a data manifold. A
learner which employs $C = C^u$ ($C^\ell = 0$) is named
\emph{Locally Preserving Projection} (LPP) \cite{He:NIPS04}.
%Belkin et al. \cite{Belkin:JMLR06}
%observed that $f^u(Z)$ can also be thought of as a norm in a Hilbert
%space \cite{Kondor:COLT03}, and is similar to a regularization term
%\cite{Scholkopf:BOOK01}; hence, thus they name this term as a
%\emph{manifold regularization}.

The parameter $\sigma$ is crucial as it controls the scale of a cost
$c^{u}_{ij}$. Hence, the choice of $\sigma$ must be sensible.
Moreover, an appropriate choice of $\sigma$ may vary across the
support of $\mathcal{P}_\textbf{x}$. Hence, the \emph{local scale}
$\sigma_i$ for each point $\textbf{x}_i$ should be used. Let
$\textbf{x}'_i$ be the $k^{th}$ nearest neighbor of $\textbf{x}_i$.
A local scale is defined as
\[
\sigma_i = \norm{\textbf{x}'_i - \textbf{x}_i},
\]
and a weight of each edge is then defined as
\begin{equation} \label{eq_localscale}
c^{u}_{ij} = \exp(\frac{\norm{\textbf{x}_i-\textbf{x}_j}^2}{\sigma_i
\sigma_j} ).
\end{equation}
Using this local scaling method is proven to be efficient in
previous experiments \cite{Perona:NIPS04} on clustering. A
specification of $k$ to define the local scale of each point is
usually more convenient than a specification of $\sigma$ since a
space of possible choices of $k$ is considerably smaller than that
of $\sigma$.

%As said in \cite{vonLuxburg:STATCOMP07}, there is no current
%agreement on the best cost matrix. At present, the task of
%constructing a similarity graph is an engineering task rather than a
%scientific task.

Instead of proposing yet another method to specify a cost matrix,
here we present a novel method which can be used to modify any
existing cost matrix. Let $Q$ and $R$ be two matrices of equal size
and have $q_{ij}$ and $r_{ij}$ as their elements. Recall that the
\emph{Hadamard product} $P$ \cite{Schott:BOOK05} between $Q$ and
$R$, $P = Q \odot R$, has elements $p_{ij} = q_{ij}r_{ij}$. In
words, the Hadamard product is a pointwise product between two
matrices. Here, we define the Hadamard $\alpha^{th}$ power operator
as
\begin{equation}
\bigodot^\alpha Q =  \overbrace{Q \odot Q \odot ... \odot Q}^{\alpha
\ times}.
\end{equation}
Given a cost matrix $C^u$ and a positive integer $\alpha$, we define
a new cost matrix $C^{u^\alpha}$ as
\begin{equation} \label{eq_hadamard_pow}
C^{u^\alpha} = \bigodot^\alpha C^u
\frac{\norm{C^u}_F}{\norm{\bigodot^\alpha C^u}_F},
\end{equation}
where $\norm{\cdot}_F$ denotes the Frobenius norm of a matrix. The
multiplication of $\frac{\norm{C^u}_F}{\norm{\bigodot^\alpha
C^u}_F}$ make $\norm{C^{u^\alpha}}_F = \norm{C^{u}}_F$. Note that if
$C^u$ is symmetric and non-negative, $C^{u^\alpha}$ still has these
properties.

The intuition of $C^{u^\alpha}$ will be explained through
experiments in Section~\ref{sect_exp} where we show that
$C^{u^\alpha}$ can further improve the quality of $C^u$ so that the
classification performance of a semi-supervised learner is
increased.

Any combinations of a label cost matrix $C^{\ell}$ of those in
previous works such as DNE, MFA and LFDA with an unlabel cost matrix
$C^{u}$ result in new SSL algorithms, and we will call the new
algorithms SS-DNE, SS-MFA and SS-LFDA.

\subsection{Non-Linear Parameterization Using the KPCA Trick} \label{sect_kpca}
By the linear parameterization, however, we can only obtain a linear
subspace defined by $A$. Learning a non-linear subspace can be
accomplished by the standard \emph{kernel trick}
\cite{Shawe:BOOK04}. However, applying the kernel trick can be
inconvenient since new mathematical formulas have to be derived and
new implementation have to be done separately from the linear
implementations. Recently, Chatpatanasiri et al. \cite{Me:Arxiv08a}
have proposed an alternative kernelization framework called
\emph{the KPCA trick}, which does not require a user to derive a new
mathematical formula or re-implement a kernelized algorithm.
Moreover, the KPCA trick framework avoids troublesome problems such
as singularity, etc.

% SAY SOMETHING ABOUT COST FUNCTION WHICH CAN be CHANGEd IF USE THE KERNEL TRICK
\subsubsection{The KPCA-Trick Algorithm}
In this section, the KPCA trick framework is extended to cover
learners implemented under our semi-supervised learning framework.
Let $k(\cdot, \cdot)$ be a PSD kernel function associated with a
non-linear function $\phi(\cdot): \Real^{d_0} \mapsto \mathcal{H}$
such that $k(\mathbf{x},\mathbf{x}') = \inner{\phi(\mathbf{x})}
{\phi(\mathbf{x}')}$ \cite{Scholkopf:BOOK01} where $\mathcal{H}$ is
a Hilbert space. Denote $\phi_i$ for $\phi(\mathbf{x}_i)$ for $i =
1, ..., \ell+u$ and $\phi'$ for $\phi(\mathbf{x}')$. The central
idea of the KPCA trick is to represent each $\phi_i$ and $\phi'$ in
a new ``finite''-dimensional space, with dimensionality bounded by
$\ell+u$, without any loss of information. Within the framework, a
new coordinate of each example is computed ``explicitly'', and each
example in the new coordinate is then used as the input of any
existing semi-supervised learner without any re-implementations.

To simplify the discussion, we assume that $\{\phi_i\}$ is linearly
independent and has its center at the origin, i.e. $\sum_i \phi_i =
0$. Since we have $n = \ell+u$ total examples, the span of
$\{\phi_i\}$ has dimensionality $n$ by our assumption. According to
\cite{Me:Arxiv08a}, each example $\phi_i$ can be represented as
$\tphi_i \in \Real^n$ with respect to a new \emph{orthonormal} basis
$\{\psi_i\}_{i=1}^n$ such that $span(\{\psi_i\}_{i=1}^n)$ is the
same as $span(\{\phi_i\}_{i=1}^n)$ without loss of any information.
More precisely, we define
\begin{equation} \label{eq_newcoord}
    \tphi_i = \Big( \inner{\phi_i}{\psi_1}, \hdots, \inner{\phi_i}{\psi_n} \Big) = \Psi^T \phi_i.
\end{equation}
where $\Psi = (\psi_1, ..., \psi_n)$. Note that although we may be
unable to numerically represent each $\psi_i$, an inner-product of
$\inner{\phi_i}{\psi_j}$ can be conveniently computed by KPCA where
each $\psi_i$ is a principal component in the feature space.
Likewise, a new test point $\phi'$ can be mapped to $\tphi' = \Psi^T
\phi'$. Consequently, the mapped data $\{\tphi_i\}$ and $\tphi'$ are
finite-dimensional and can be explicitly computed.

The KPCA-trick algorithm consisting of three simple steps is shown
in Figure~\ref{fig_KPCA}. All semi-supervised learners can be
kernelized by this simple algorithm. In the algorithm, we denote a
semi-supervised learner by \textsf{ssl} which outputs the best
linear map $A^*$.

\begin{figure}[h]
\begin{center}
\vskip -0.0in \fbox{
\begin{minipage}{11cm}
  \textbf{Input:} 1. training examples: $\{(\textbf{x}_1,y_1), ..., (\textbf{x}_\ell,y_\ell),\textbf{x}_{\ell+1}, ..., \textbf{x}_{\ell+u}\}$\\
  \hspace*{1.13cm} 2. a new example: $\textbf{x}'$\\
  \hspace*{1.13cm} 3. a kernel function: $k(\cdot,\cdot)$\\
  \hspace*{1.13cm} 4. a linear semi-supervised learning algorithm: \textsf{ssl} (see Figure~\ref{fig_ssl})\\
\textbf{Algorithm:}\\
    (1) Apply \textsf{kpca}($k,\{\textbf{x}_i\}_{i=1}^{\ell+u}$, $\textbf{x}'$) such that $\{\textbf{x}_i\} \mapsto \{\tphi_i\}$
    and $\textbf{x}' \mapsto \tphi'$.\\
    (2) Apply \textsf{ssl} with new inputs $\{(\tphi_1,y_1), ..., (\tphi_\ell,y_\ell),\tphi_{\ell+1}, ..., \tphi_{\ell+u}\}$\\
    \hspace*{0.47cm} to achieve $A^*$.\\
    (3) Perform kNN based on the distance $\norm{A^*\tphi_i - A^*\tphi'}$.\\
\end{minipage}
} \vskip -0.1in \caption{The KPCA-trick algorithm for
semi-supervised learning.} \label{fig_KPCA}
\end{center}
\vskip -0.1in
\end{figure}

\subsection{Remarks} \label{sect_remark}
\noindent 1. The main optimization problem shown in
Eq.\eqref{eq_linear2} can be restated as follows:
\cite{Fukunaga:Book90}
\begin{equation*} \label{eq_linear5}
  \underset{A \in \Real^{d \times d_0}}{\operatorname{\argmin}}\, \mbox{trace}\Big((A B A^T)^{-1}AX(D-C)X^TA^T\Big).
\end{equation*}
Within this formulation, the corresponding optimal solution is
invariant under a non-singular linear transformation; i.e., if $A^*$
is an optimal solution, then $T A^*$ is also an optimal solution for
any non-singular $T \in \Real^{d \times d}$
\cite[pp.447]{Fukunaga:Book90}. We note that four choices of $T$
which assign a weight to each new axis are natural: (1)~$T = I$,
(2)~$T$ is a diagonal matrix with $T_{ii} =
\frac{1}{\norm{\textbf{a}^{(i)}} }$, i.e. $T$ normalizes each axis
to be equally important, (3)~$T$ is a diagonal matrix with $T_{ii} =
\sqrt{\lambda_i}$ as $\sqrt{\lambda_i}$ determines how well each
axis $\textbf{a}^{(i)}$ fits the objective function $\textbf{a}^{(i)
T} X(D-C)X^T \textbf{a}^{(i)}$, and (4)~$T$ is a diagonal matrix
with $T_{ii} = \frac{\sqrt{\lambda_i}}{ \norm{\textbf{a}^{(i)}} }$,
i.e. a combination of (2) and (3).\\

%\footnote{LFDA implementation (http://sugiyama-www.cs.titech
%.ac.jp/$\sim$sugi/software/) also allows similar choices of $T$.}

\noindent 2. The matrices $B$ defined in Subsection~\ref{sect_cost}
of the two algorithms, SS-MFA and SS-LFDA,  are guarantee to be
positive semidefinite (PSD) but may not be positive definite (PD),
i.e., $B$ may not be full-rank. In this case, $B$ is singular and we
cannot immediately apply Eq.\eqref{eq_framework3} to solve the
optimization problems. One common way to solve this difficulty is to
use $(B + \epsilon I)$, for some value of $\epsilon$, which is now
guaranteed to be full-rank instead of $B$ in Eq.\eqref{eq_linear2}.
Since $\epsilon$ acts in a role of regularizer, it makes sense to
set $\epsilon = \gamma$, the regularization parameter specified in
Section~\ref{sect_cost}. Similar settings of $\epsilon$ have also
been used by some existing algorithms, e.g.
\cite{Friedman:JASA89,Sugiyama:pakdd08}.
%Nevertheless, an
%answer of a problem depends on $\epsilon$, and a process of tuning
%the value of $\epsilon$ is needed.

Also, in a small sample size problem where $X(D-C)X^T$ is not
full-rank, the obtained matrix $A^*$ (or some columns of $A^*$) lie
in the null space of $X(D-C)X^T$. Although this matrix does optimize
our optimization problem, it usually overfits the given data. One
possible solution to this problem is to apply PCA to the given data
in the first place \cite{Belhumeur:PAMI97} so that the resulted data
have dimensionality less than or equal to the rank of $X(D-C)X^T$.
Note that in our KPCA trick framework this pre-process is
automatically accomplished as KPCA has to be applied to a learner as
shown in Figure~\ref{fig_KPCA}.\\

\section{Related Work: Connection and Improvement}
\label{sect_related} As we already described in
Section~\ref{sect_cost}, our framework generalizes various existing
supervised and unsupervised manifold learners
\cite{Sugiyama:JMLR07,Yan:PAMI07,Wei:ICML07,Cai:CVPR07,Hoi:CVPR06,Chen:CVPR05,Cheng:ICIP04,vonLuxburg:STATCOMP07,He:NIPS04,Perona:NIPS04}.
The KPCA trick is new in the field of semi-supervised learning.

There are some supervised manifold learners which cannot be
represented in our framework
\cite{Goldberger:NIPS05,Globerson:NIPS06,Weinberger:NIPS06,Yang:AAAI06,Torresani:NIPS07}
because their cost functions are not linear with respect to
distances among examples. Extension of these algorithms to handle
semi-supervised learning problems is an interesting future work.

Yang et al. \cite{Yang:ICML2006} present another semi-supervised
learning framework which solves entirely different problems to
problems considered in this paper. They propose to extend
unsupervised algorithms such as ISOMAP \cite{Tenenbaum:Science00}
and Laplacian Eigenmap \cite[Chapter 16]{Semi:BOOK06} to cases to
which information about exact locations of some points is available.

%The idea of manifold regularization $f^u(\cdot)$ is first used by
%Belkin et al. \cite{Belkin:JMLR06}. In their work, they apply
%manifold regularization to support vector machines and regularized
%least square. The target dimensionality $d$ is limited to 1 in both
%cases. Here, we apply manifold regularization to dimensionality
%reduction algorithms for any $d \le d_0$.
%Their representer theorem
%is also not valid for the cases of $d > 1$.

To the best of our knowledge, there are currently two existing
semi-supervised dimensionality reduction frameworks in literatures
which have similar goal to ours; both are very recently proposed.
Here, we subsequently show that these frameworks can be restated as
\emph{special cases} of our framework.

%\subsection{Yan et al. \cite{Yan:PAMI07}}
%Our framework is also inspired from \emph{graph embedding} framework
%of Yan et al. \cite{Yan:PAMI07}. However, their framework considers
%only supervised and unsupervised problems. It can be said that we
%extend their framework to handle SSL problems.

\subsection{Sugiyama et al. \cite{Sugiyama:pakdd08}}
\label{sect_self} Sugiyama et al. \cite{Sugiyama:pakdd08} extends
the LFDA algorithm to handle a semi-supervised learning problem by
adding the PCA objective function $f^{PCA}(A)$ into the objective
function $f^\ell(A)$ of LFDA described in Section~\ref{sect_cost}.
To describe Sugiyama et al.'s algorithm, namely `SELF', without loss
of generality, we assume that training data are centered at the
origin, i.e. $\sum_{i=1}^n \textbf{x}_i = 0$, and then we can write
$f^{PCA}(A) = -\sum_{i=1}^n \norm{A\textbf{x}_i}^2$. Sugiyama et al.
propose to solve the following problem:
\begin{equation}
A^* = \underset{A B A^T = I}{\operatorname{\argmin}}\,
\left(\sum_{i,j=1}^\ell c^\ell_{ij}\norm{A\textbf{x}_i -
A\textbf{x}_j}^2 - \gamma \sum_{i=1}^n \norm{A\textbf{x}_i}^2
\right)
\end{equation}
Interestingly, it can be shown that this formulation can be
formulated in our framework with unlabel cost $c^{u}_{ij}$ being
\emph{negative}, and hence our framework subsumes SELF. To see this,
let $c^{u}_{ij} = -1/2n$, for all $i,j = 1, ..., n$. Then, the
objective $f^u(A)$ is equivalent to $f^{PCA}(A)$: \mathindent=0mm
\begin{align*}
f^u(A) = \sum_{i,j=1}^n -\frac{1}{2n} \norm{A\textbf{x}_i -
A\textbf{x}_j}^2
 &= -\frac{1}{2n} \sum_{i,j=1}^n \inner{A\textbf{x}_i - A\textbf{x}_j}{A\textbf{x}_i - A\textbf{x}_j}\\
 &= -\frac{1}{2n} \left( 2 \sum_{i,j=1}^n \inner{A\textbf{x}_i}{A\textbf{x}_i}
     - 2 \sum_{i,j=1}^n \inner{A\textbf{x}_i}{A\textbf{x}_j}\right) \\
 &= -\frac{1}{2n} \left( 2n \sum_{i=1}^n \norm{A\textbf{x}_i}^2
     - 2  \inner{A \sum_{i=1}^n\textbf{x}_i}{A\sum_{j=1}^n \textbf{x}_j }\right) \\
 &= f^{PCA}(A),
\end{align*} \mathindent=7mm
where we use the fact that $\sum_{i=1}^n \textbf{x}_i = 0$. This
proves that SELF is a special case of our framework.

Note that the use of negative unlabel costs $c^{u}_{ij} = -1/2n$
results in an algorithm which tries to preserve a \emph{global}
structure of the input data and does not convey the manifold
assumption where only a \emph{local} structure should be preserved.
Therefore, when the input unlabeled data lie in a complicated
manifold, it is not appropriate to apply $f^{u}(A) = f^{PCA}(A)$.

\subsection{Song et al. \cite{Song:PR08}}
%Recently, a, b and c propose ssl algorithms
%\cite{Song:PR08,Sugiyama:ssl08}. Interestingly, all those algorithms
%can be casted into our framework.

%thus, their algorithms are inferior to SS-LFDA, SS-MFA and SS-DNE
%described in this paper

%, and thus the work of Song et al. is a special case of our
%framework

Song et al. propose to extend FDA and another algorithm named
\emph{maximum margin criterion} (MMC) \cite{Haifeng:IEEENN06} to
handle a semi-supervised learning problem. Their idea of
semi-supervised learning extension is similar to ours as they add
the term $f^u(\cdot)$ into the objective of FDA and MMC (hence, we
call them, SS-FDA and SS-MMC, respectively). However, SS-FDA and
SS-MMC cannot handle problems where data of each class form a
manifold or several clusters as shown in Figure~\ref{fig_nearby}
because SS-FDA and SS-MMC satisfy the condition (1) but not (1*). In
fact, SS-FDA and SS-MMC can both be restated as instances of our
framework. To see this, we note that the optimization problem of
SS-MMC can be stated as
\begin{equation}
  A^* = \underset{A A^T = I}{\operatorname{\argmin}}\, \ \
        \gamma' \mbox{trace}(A  S_w  A^T) - \mbox{trace}(A  S_b  A^T) + \gamma f^u(A),
\end{equation}
where $S_b$ and $S_w$ are standard between-class and within-class
scatter matrices, respectively \cite{Fukunaga:Book90}:
\[
 S_w = \sum_{i=1}^c \sum_{j | y_j = i}
  (\textbf{x}_j - \jbf{\mu}_i)(\textbf{x}_j - \jbf{\mu}_i)^T
  \ \ \mbox{and} \ \
 S_b = \sum_{i=1}^c
  (\jbf{\mu} - \jbf{\mu}_i)(\jbf{\mu} - \jbf{\mu}_i)^T,
\]
where $\jbf{\mu}= \frac{1}{n}\sum_{i=1}^n \textbf{x}_i$,
$\jbf{\mu}_i = \frac{1}{n_i}\sum_{i=1}^{n_i} \textbf{x}_i$ and $n_i$
is the number of examples in the $i^{th}$ class. It can be checked
that $\mbox{trace}(A S_w A^T) = \sum_{i,j=1}^\ell
c^w_{ij}\norm{A\textbf{x}_i - A\textbf{x}_j}^2$ and $\mbox{trace}(A
S_b  A^T) = \sum_{i,j=1}^\ell c^b_{ij}\norm{A\textbf{x}_i -
A\textbf{x}_j}^2$ where
\begin{equation*}
c^{b}_{ij} =
\begin{cases}
   (\frac{1}{n_k} - \frac{1}{n}), & \mbox{if } y_i = y_j = k,\\
   -\frac{1}{n}, & \mbox{otherwise},
\end{cases} \ \  \mbox{and} \ \
c^{w}_{ij} =
\begin{cases}
   \frac{1}{n_k}, & \mbox{if } y_i = y_j = k, \\
   0, & \mbox{otherwise}.
\end{cases}
\end{equation*}
%\begin{equation*}
%\end{equation*}.
Hence, by setting $c^{\ell}_{ij} = \gamma' c^{w}_{ij} - c^{b}_{ij}$
we finish our proof that SS-MMC is a special case of our framework.
The proof that SS-FDA is in our framework is similar to that of
SS-MMC.

\subsection{Improvement over Previous Frameworks}
\begin{figure}[t]
\begin{center}
\vskip -0.2in \setlength{\epsfxsize}{3.75in}
\centerline{\epsfbox{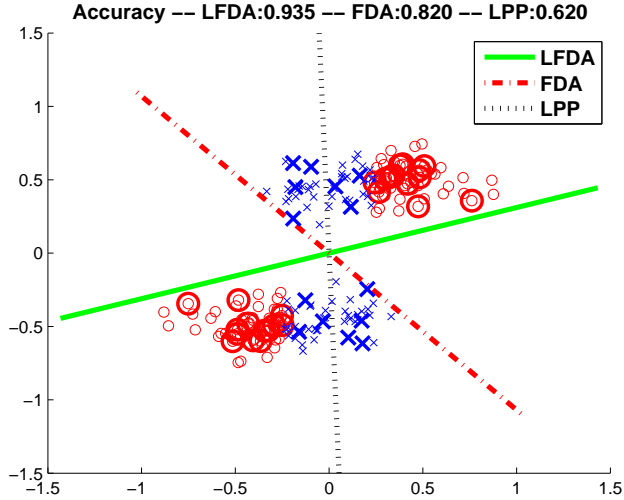}} \vskip -0.2in \caption{The
first toy example. The projection axes of three algorithms, namely
FDA, LFDA, and LPP, are presented. Big circles and big crosses
denote labeled examples while small circles and small crosses denote
unlabeled examples. Their percentage accuracy over the unlabeled
examples are shown on the top.} \label{fig_toy} \vskip -0.45in
\end{center}
\end{figure}

\begin{figure}[t]
\begin{center}
\vskip -0.2in \setlength{\epsfxsize}{3.75in}
\centerline{\epsfbox{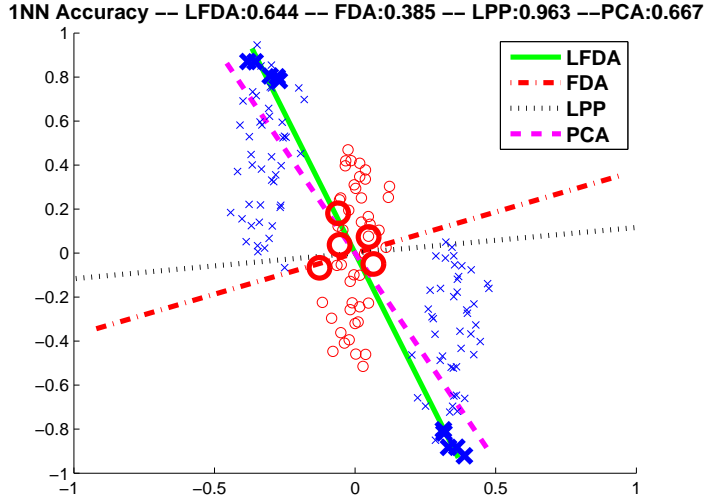}} \vskip -0.2in
\caption{The second toy example consisting of three clusters of two
classes.} \label{fig_toy2} \vskip -0.45in
\end{center}
\end{figure}

In this section, we explain why SELF and SS-FDA proposed by Sugiyama
et al. \cite{Sugiyama:pakdd08} and Song et al. \cite{Song:PR08}
described above are \emph{not} enough to solve some semi-supervised
learning problems, even simple ones shown in Figure~\ref{fig_toy}
and Figure~\ref{fig_toy2}.

In Figure~\ref{fig_toy}, three dimensionality reduction algorithms,
FDA, LFDA and LPP are performed on this dataset. Because of
multi-modality, FDA cannot find an appropriate projection. Since the
two clusters do not contain data of the same class, LPP which tries
to preserve the structure of the two clusters also fails. In this
case, only LFDA can find a proper projection since it can cope with
multi-modality and can take into account the labeled examples. Note
that since SS-FDA is a linearly combined algorithm of FDA and LPP,
it can only find a projection lying in between the projections
discovered by FDA and LPP, and in this case SS-FDA cannot find an
efficient projection, unlike LFDA and, of course, SS-LFDA derived
from our framework.

A similar argument can be given to warn an uncareful use of SELF in
some situations. In Figure~\ref{fig_toy2}, four dimensionality
reduction algorithms, FDA, PCA, LFDA and LPP are performed on this
dataset. Because of multi-modality, FDA and PCA cannot find an
appropriate projection. Also, since there are only a few labeled
examples, LFDA fails to find a good projection as well. In this
case, only LPP can find a proper projection since it can cope with
multi-modality and can take the unlabeled examples into account.
Note that since SELF is a linearly combined algorithm of LFDA and
PCA, it can only find a projection lying in between the projections
discovered by LFDA and PCA, and in this case SELF cannot find a
correct projection, unlike a semi-supervised learner like SS-LFDA
derived from our framework which, as explained in
Section~\ref{sect_cost},  employs the LPP cost function as its
$C^u$.

\begin{figure*}[t]  \begin{center} \vskip -0.2in \hbox{\hskip -0.8in
\setlength{\epsfxsize}{3.15in} \epsfbox{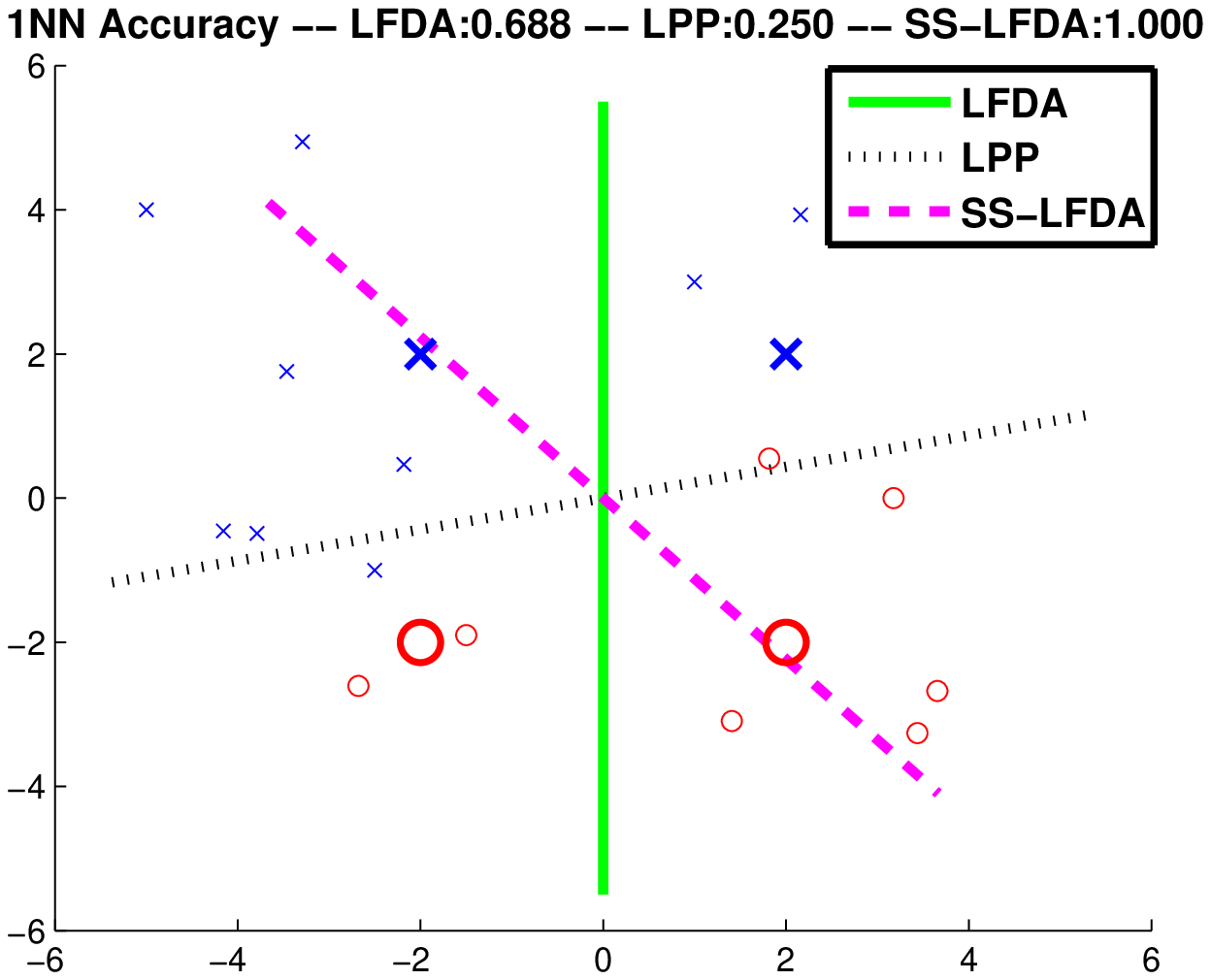}
\setlength{\epsfxsize}{3.15in} \epsfbox{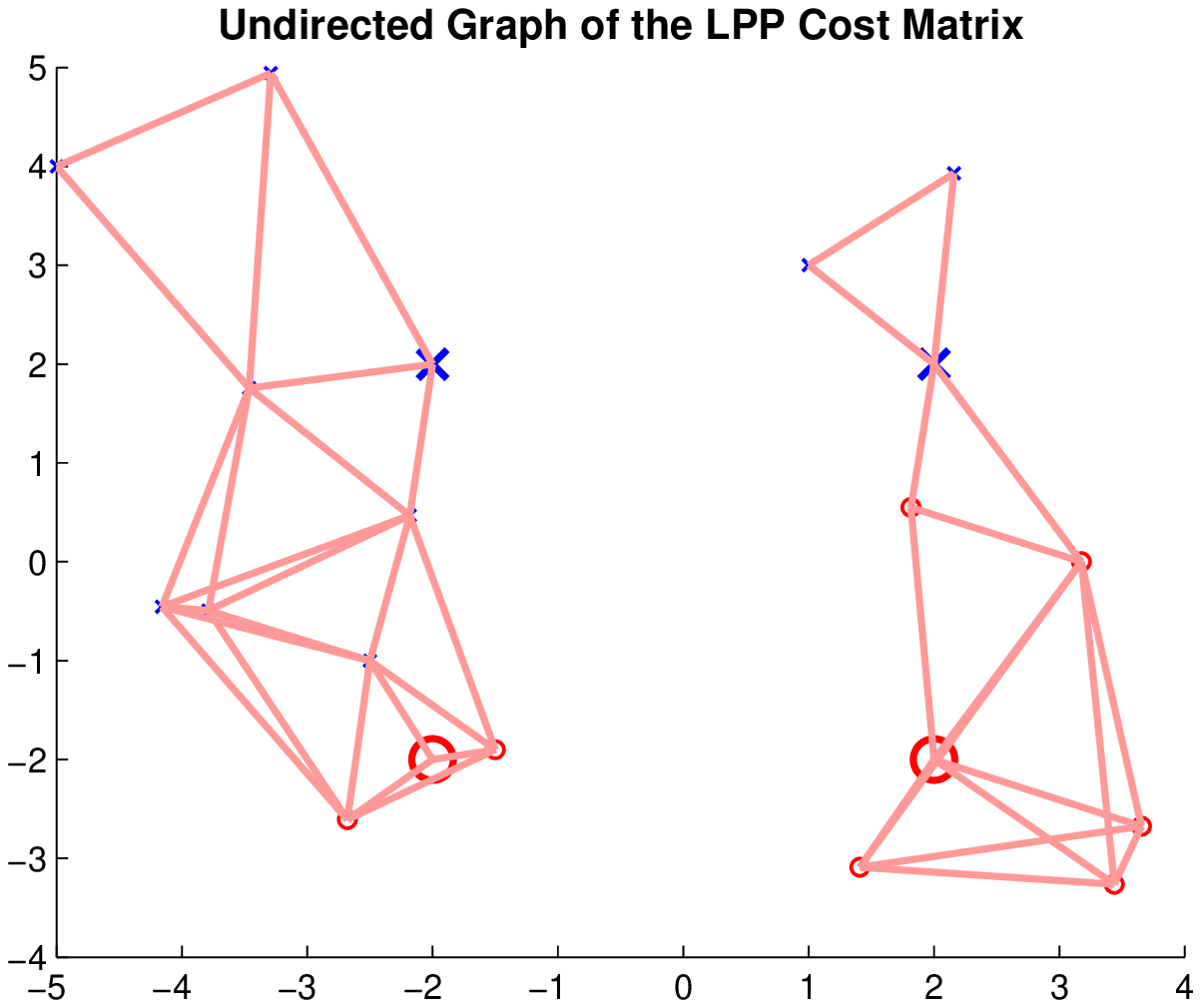} }
\end{center}
\vskip -0.5in \caption{ (Left) The third toy example where only a
semi-supervised learner is able to find a good projection. (Right)
An undirected graph corresponding to the values of $C^u$ used by LPP
and SS-LFDA. In this figure, a pair of examples $i$ and $j$ has a
link if and only if $c^u_{ij} > 0.1$. This graph explains why LPP
projects the data in the axis shown in the left figure; LPP, which
does not apply the label information, tries to choose a projection
axis which squeezes the two clusters as much as possible. Note that
we apply a local-scaling method, Eq.\eqref{eq_localscale}, to
specify $C^u$.} \label{fig_toy3} \vskip -0.0in
\end{figure*}

Since a semi-supervised manifold learner derived from our framework
can be intuitively thought of as a combination of a supervised
learner and an unsupervised learner. One may misunderstand that a
semi-supervised learner cannot discover a good subspace if neither
is a supervised learner nor an unsupervised learner able to discover
a good subspace. The above two toy examples may also mislead the
readers in that way. In fact, that intuition is incorrect. Here, we
give another toy example shown in Figure~\ref{fig_toy3} where only a
semi-supervised learner is able to discover a good subspace but
neither is its supervised and unsupervised counterparts.
Intuitively, a semi-supervised learner is able to exploit useful
information from both labeled and unlabeled examples.

\section{Experiments} \label{sect_exp}

In this section, classification performances of algorithms derived
from our framework are demonstrated. We try to use a similar
experimental setting as those in previous works \cite[Chapter
21]{Sugiyama:pakdd08,Semi:BOOK06} so that our results can be
compared to them.

\subsection{Experimental Setting} \label{sect_exp_setting}
In all experiments, two semi-supervised learners, SS-LFDA and
SS-DNE, derived from our framework are compared to relevant existing
algorithms, PCA, LPP*, LFDA, DNE and SELF \cite{Sugiyama:pakdd08}.
In contrast to the standard LPP which does not apply the Hadamard
power operator explained in Section~\ref{sect_cost}, we denote LPP*
as a variant of LPP applying the Hadamard power operator.

Non-linear semi-supervised manifold learning is also experimented by
applying the KPCA trick algorithm illustrated in
Figure~\ref{fig_KPCA}. Since it is not our intention to apply the
``best'' kernel \emph{but} to compare efficiency between a
``semi-supervised'' kernel learner and its base ``supervised'' (and
``unsupervised'') kernel learners, we simply apply the
$2^{nd}$-degree polynomial kernel $k(\textbf{x},\textbf{x}') =
\inner{\textbf{x}}{\textbf{x}'}^2$ to the kernel algorithms in all
experiments.

By using the nearest neighbor algorithm on their discovered
subspaces, classification performances of the experimented learners
are measured on five standard datasets shown on
Table~\ref{table_data}, the first two datasets are obtained from the
UCI repository \cite{data:UCI}, the next two datasets mainly
designed for testing a semi-supervised learner are obtained from
\textsf{http://www.kyb.tuebingen.mpg.de/ssl-book/benchmarks.html}
\cite[Chapter 21]{Semi:BOOK06}. The final dataset, \emph{extended
Yale B} \cite{Kriegman:PAMI01}, is a standard dataset of a face
recognition task. The classification performance of each algorithm
is measured by the average test accuracy over 25 realizations of
randomly splitting each dataset into training and testing subsets.
%i.e., 20 Out-of-bag
%experiments \cite{Breiman:AnStat98,Breiman:mlj96}

\begin{table}[t] \caption{Details of each dataset: $d_0, c, \ell, u$ and $t$ denote
the numbers of input features, classes, labeled examples, unlabeled
examples and testing examples, respectively. `*' denotes the
transductive setting used in small datasets, where all examples
which are not labeled are given as unlabeled examples and used as
testing examples as well. $d$, determined by using prior knowledge,
denotes the target dimensionality for each dataset. ``\textsc{Good
Neighbors}'' denotes a quantity which measures a goodness of
unlabeled data for each dataset.} \label{table_data}
%\vskip 0.10in
\begin{center}
\begin{small}
\begin{sc}
\begin{tabular}{l|ccccc|c|cc}
\hline
%\abovespace\belowspace
Name & $d_0$ & $c$ & $\ell+u+t$ & $\ell$ & $u$ & $d$ & \multicolumn{2}{|c}{Good Neighbors}\\
 & & & & & & & linear & kernel\\
\hline
%\abovespace
Ionosphere     & 34 & 2 & 351  & 10/100 & * &      2 & 0.866  & 0.843\\
Balance        &  4 & 3 & 625  & 10/100 & 300 &      1 & 0.780  & 0.760\\
%G241c          &241 & 2 & 1500 & 10/100 & 300    & 1 & 0.636  & 0.503\\
BCI            &117 & 2 & 400 & 10/100 & *         & 2& 0.575  & 0.593  \\
Usps           &241 & 2 & 1500 & 10/100 & 300    & 10& 0.969  & 0.971\\
M-Eyale        &504 & 5 & 320  & 20/100 & * &      10& 0.878 &  0.850\\
%\belowspace
\hline
\end{tabular}
\end{sc}
\end{small}
\end{center}
%\vskip -0.2in
\end{table}

Three parameters are needed to be tuned in order to apply a
semi-supervised learner derived from our framework (see
Section~\ref{sect_cost}): $\gamma$, the regularizer, $\alpha$, the
degree of the Hadamard power operator and $k$, the $k^{th}$-nearest
neighbor parameter needed to construct the cost matrices. To make
our learners satisfy the condition (1*) described in
Section~\ref{sect_cost}, it is clear that $k$ should be small
compared to $n_c$, the number of training examples of class $c$.
From our experience, we found that semi-supervised learners are
quite insensitive to various small values of $k$. Therefore, in all
our experiments, we simply set $k = \mbox{min}(3,n_c)$ so that only
two parameters, $\gamma$ and $\alpha$, are needed to be tuned. We
tune these two parameters via cross validation. Note that only
$\alpha$ is needed to be tuned for LPP* and only $\gamma$ is needed
to be tuned for SELF.

The `\textsc{Good Neighbors}' score shown in Table~\ref{table_data}
is due to Sugiyama et al. \cite{Sugiyama:pakdd08}. The score is
simply defined as a training accuracy of the nearest neighbor
algorithm when \emph{all available data are labeled and are given to
the algorithm}. Note that this score is not used by a dimensionality
reduction algorithm. It just clarifies a usefulness of unlabeled
examples of each dataset to the readers. Intuitively, if a dataset
gets a high score, unlabeled examples should be useful since it
indicates that each pair of examples having a high penalty cost
$c^u_{ij}$ should belong to the same class. Note that on
Table~\ref{table_data} there are two scores for each dataset:
\textsc{linear} is a score on a given input space while
\textsc{kernel} measures a score on a feature space corresponding to
the $2^{nd}$-degree polynomial kernel.
% we should actually generalize this score to cover the k-nearest neighbor graph;
% as now this score is intuitive only k=1;

\subsection{Numerical Results}
Numerical results are shown in Table~\ref{table_expl10} for the case
of $\ell=10$ (except \textsc{M-Eyale} where $\ell=20$) and
Table~\ref{table_expl100} for the case of $\ell=100$. In
experiments, SS-DNE and SS-LFDA are compared their classification
performances to their unsupervised and supervised counterparts: LPP*
and DNE for SS-DNE, and LPP* and LFDA for SS-LFDA. SELF is also
compared to SS-LFDA as they are related semi-supervised learners
originated from LFDA. Our two algorithms will be highlighted if they
are superior to their counterpart opponents.

From the results, our two algorithms, SS-LFDA and SS-DNE, outperform
all their opponents in 32 out of 40 comparisons: in the first
setting of small $\ell$ (Table~\ref{table_expl10}), our algorithms
outperform the opponents in 18 out of 20 comparisons while in the
second setting of large $\ell$ (Table~\ref{table_expl100}), our
algorithms outperform the opponents in 14 out of 20 comparisons.
Consequently, our framework offers a semi-supervised learner which
consistently improves its base supervised and unsupervised learners.

Note that as the number of labeled examples increases, usefulness of
unlabeled examples decreases. We will subsequently discuss and
analyze the results of each dataset in details in the next
subsections.

\begin{table}[t] \vskip -0.15in \caption{Percentage accuracies of SS-DNE and SS-LFDA derived from our framework compared to existing algorithms ($\ell = 10$, except \textsc{M-Eyale} where $\ell = 20$).
SS-LFDA and SS-DNE are highlighted when they outperform their
opponents (LPP* and DNE for SS-DNE, and LPP*, LFDA and SELF for
SS-LFDA). Superscripts indicate \%-confidence levels of the
one-tailed paired t-test for differences in accuracies between our
algorithms and their best opponents. No superscripts denote
confidence levels which below 80\%.} \label{table_expl10}
\begin{center}
\begin{small}\hskip -0.20in
\begin{sc}
\vskip -0.20in
\begin{tabular}{l|ccccc|cc}
\hline
%\abovespace\belowspace
Linear &          PCA &        LPP* &         DNE &        LFDA &        SELF &     SS-DNE &   SS-LFDA\\
\hline
%\abovespace
Ionosphere  & 71$\pm$1.2 & 82$\pm$1.3  & 70$\pm$1.2 & 71$\pm$1.1 & 70$\pm$1.5 & 75$\pm$1.0  & 78.1$\pm$.9\\
Balance     & 49$\pm$1.9 & 61$\pm$1.9  & 63$\pm$2.2 & 70$\pm$2.2 & 69$\pm$2.3 & \textbf{71$\pm$1.8}$^{99}$  & \textbf{73$\pm$2.3}$^{80}$\\
BCI         & 49.8$\pm$.6& 53.4$\pm$.3 & 51.3$\pm$.6& 52.6$\pm$.5 & 52.1$\pm$.5 & \textbf{57.1$\pm$.6}$^{99}$ & \textbf{55.2$\pm$.3}$^{99}$ \\
Usps        & 79$\pm$1.2 &  74$\pm$1.0  &  79.6$\pm$.6  &  80.6$\pm$.9  &  81.7$\pm$.8 &  \textbf{81.8$\pm$.5}$^{99}$ & \textbf{83.0$\pm$.5}$^{90}$\\
M-Eyale     & 44.6$\pm$.7 & 67$\pm$1.1 & 66$\pm$1.2 & 71.6$\pm$1.0 & 67.2$\pm$.8 & \textbf{76.9$\pm$.8}$^{99}$ & \textbf{75.7$\pm$.9}$^{99}$ \\
%\belowspace
\hline \hline
Kernel &          PCA &        LPP* &         DNE &        LFDA &        SELF &     SS-DNE &   SS-LFDA\\
\hline
%\abovespace
Ionosphere  & 70$\pm$1.8 & 83.2$\pm$.9  & 70$\pm$1.6 & 71$\pm$1.3 & 74$\pm$1.5 & \textbf{87.2$\pm$.9}$^{99}$  & \textbf{88$\pm$1.0}$^{99}$\\
Balance     & 41.7$\pm$.8 & 47.9$\pm$.9  & 62$\pm$2.5 & 66$\pm$2.0 & 60$\pm$2.8 & \textbf{66$\pm$1.8}$^{80}$  & \textbf{69$\pm$1.9}$^{80}$\\
BCI         & 49.7$\pm$.3  & 53.7$\pm$.3 & 50.1 $\pm$.4 & 50.3$\pm$.6 & 50.5$\pm$.4 & \textbf{53.8$\pm$.3} & \textbf{54.1$\pm$.3}$^{80}$ \\
Usps        & 77$\pm$1.1 & 76$\pm$1.1 & 79.9$\pm$.5 & 80.3$\pm$.8 & 80.9$\pm$.8 &  \textbf{82.0$\pm$.4}$^{99}$  & \textbf{83.7$\pm$.6}$^{99}$\\
M-Eyale     & 42.1$\pm$.9& 63.2$\pm$.7 & 58.0$\pm$.9 & 60.3$\pm$.8 & 58.8$\pm$.7 & \textbf{69.9$\pm$.7}$^{99}$ & \textbf{73.2$\pm$.8}$^{99}$ \\
\hline
\end{tabular}
\end{sc}
\end{small}
\end{center}
\vskip -0.0in
\end{table}

\begin{table}[t] \vskip -0.15in \caption{Percentage accuracies of SS-DNE and SS-LFDA compared to existing algorithms ($\ell = 100$).}
\label{table_expl100}
\begin{center}
\begin{small}\hskip -0.20in
\begin{sc}
\vskip -0.20in
\begin{tabular}{l|ccccc|cc}
\hline
%\abovespace\belowspace
Linear &          PCA &        LPP* &         DNE &        LFDA &        SELF &     SS-DNE &   SS-LFDA\\
\hline
%\abovespace
Ionosphere  & 72.8$\pm$.6 & 83.7$\pm$.6 & 77.9$\pm$.7 & 74$\pm$1.0 & 77.8$\pm$.5 & \textbf{84.5$\pm$.6}$^{80}$ & \textbf{84.9$\pm$.4}$^{95}$\\
Balance     & 57$\pm$2.2 & 80$\pm$1.3   & 86.4$\pm$.5 & 87.9$\pm$.3 & 87.2$\pm$.4 & \textbf{88.2$\pm$.5}$^{99}$  & 86.3$\pm$.6\\
BCI         & 49.5$\pm$.5 & 54.9$\pm$.5 & 53.1$\pm$.7 & 67.9$\pm$.5 & 67.6$\pm$.6 &  \textbf{63.1$\pm$.5}$^{99}$   & 67.5$\pm$.6 \\
Usps        & 91.4$\pm$.3 & 75.7$\pm$.3 & 91.1$\pm$.3 & 89.3$\pm$.4 & 92.2$\pm$.3 & \textbf{92.2$\pm$.4}$^{95}$ & 91.6$\pm$.3 \\
M-Eyale     & 69.4$\pm$.4 & 84.1$\pm$.4 & 92.3$\pm$.4 & 95.4$\pm$.3 & 94.3$\pm$.2 & \textbf{93.5$\pm$.4}$^{95}$ & \textbf{95.7$\pm$.2} \\
%\belowspace
\hline \hline
Kernel &          PCA &        LPP* &         DNE &        LFDA &        SELF &     SS-DNE &   SS-LFDA\\
\hline
%\abovespace
Ionosphere  & 79.8$\pm$.4 & 89.7$\pm$.5 & 78.7$\pm$.9 & 81.3$\pm$.7 & 81.1$\pm$.5 & \textbf{93.6$\pm$.2}$^{99}$  & \textbf{93.7$\pm$.3}$^{99}$\\
Balance     & 42.5$\pm$.3 & 46.9$\pm$.5 & 84.0$\pm$.7 & 87.8$\pm$.7 & 79$\pm$1.6 & \textbf{86.5$\pm$.7}$^{99}$  & 87.7$\pm$.9\\
BCI         & 49.7$\pm$.5 & 54.5$\pm$.4 & 51.6$\pm$.6 &  51.0$\pm$.8 & 52.4$\pm$.6 & \textbf{57.6$\pm$.2}$^{99}$ & \textbf{57.0$\pm$.4}$^{99}$ \\
Usps        & 91.1$\pm$.3 & 81.5$\pm$.6 & 91.4$\pm$.4 & 91.2$\pm$.4 & 92.7$\pm$.3 & \textbf{92.3$\pm$.3}$^{95}$ & 91.9$\pm$.3 \\
M-Eyale     & 66.3$\pm$.3 & 81.9$\pm$.5 & 91.2$\pm$.3 & 89.1$\pm$.5 & 85.8$\pm$.6 & 91.2$\pm$.3 & \textbf{94.3$\pm$.3}$^{99}$ \\
\hline
\end{tabular}
\end{sc}
\end{small}
\end{center}
\vskip -0.2in
\end{table}

%\subsection{Discussion and Analysis of Experiments}
\subsubsection{Ionosphere}
\textsc{Ionosphere} is a real-world dataset of radar pulses passing
through the ionosphere which were collected by a system in Goose
Bay, Labrador.  The targets were free electrons in the ionosphere.
``Good'' radar returns are those showing evidence of some type of
structure in the ionosphere. ``Bad'' returns are those that do not.
Since we do not know the true decision boundary of
\textsc{Ionosphere}, we simply set the target dimensionality $d = c
= 2$. It can be observed that non-linearization does improve the
classification performance of all algorithms.

\begin{figure}[t]
\begin{center}
\vskip -0.2in \setlength{\epsfxsize}{5in}
\centerline{\epsfbox{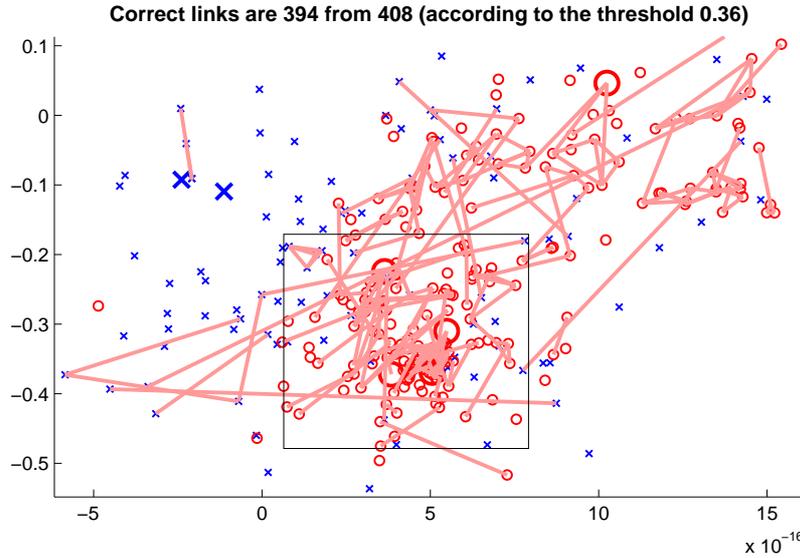}} \vskip -0.2in
\caption{The undirected graph corresponding to $C^u$ constructed on
\textsc{Ionosphere}. Each link corresponds to a pair of nearby
examples having $c^u_{ij} \ge 0.36$. The number `0.36' is just
chosen for visualizability.} \label{fig_simiarlity_ionos1} \vskip
-0.25in
\end{center}
\end{figure}

\begin{figure}[t]
\begin{center}
\vskip -0.2in \setlength{\epsfxsize}{4.5in}
\centerline{\epsfbox{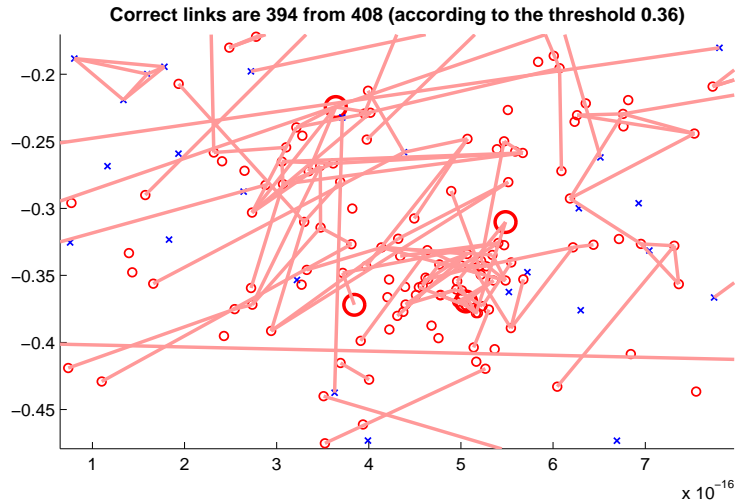}} \vskip -0.2in
\caption{Zoom-in on the square area of
Figure~\ref{fig_simiarlity_ionos2}.} \label{fig_simiarlity_ionos2}
\vskip -0.05in
\end{center}
\end{figure}

\begin{figure*}[h]  \begin{center} \vskip -0.2in \hbox{\hskip -0.8in
\setlength{\epsfxsize}{3.15in} \epsfbox{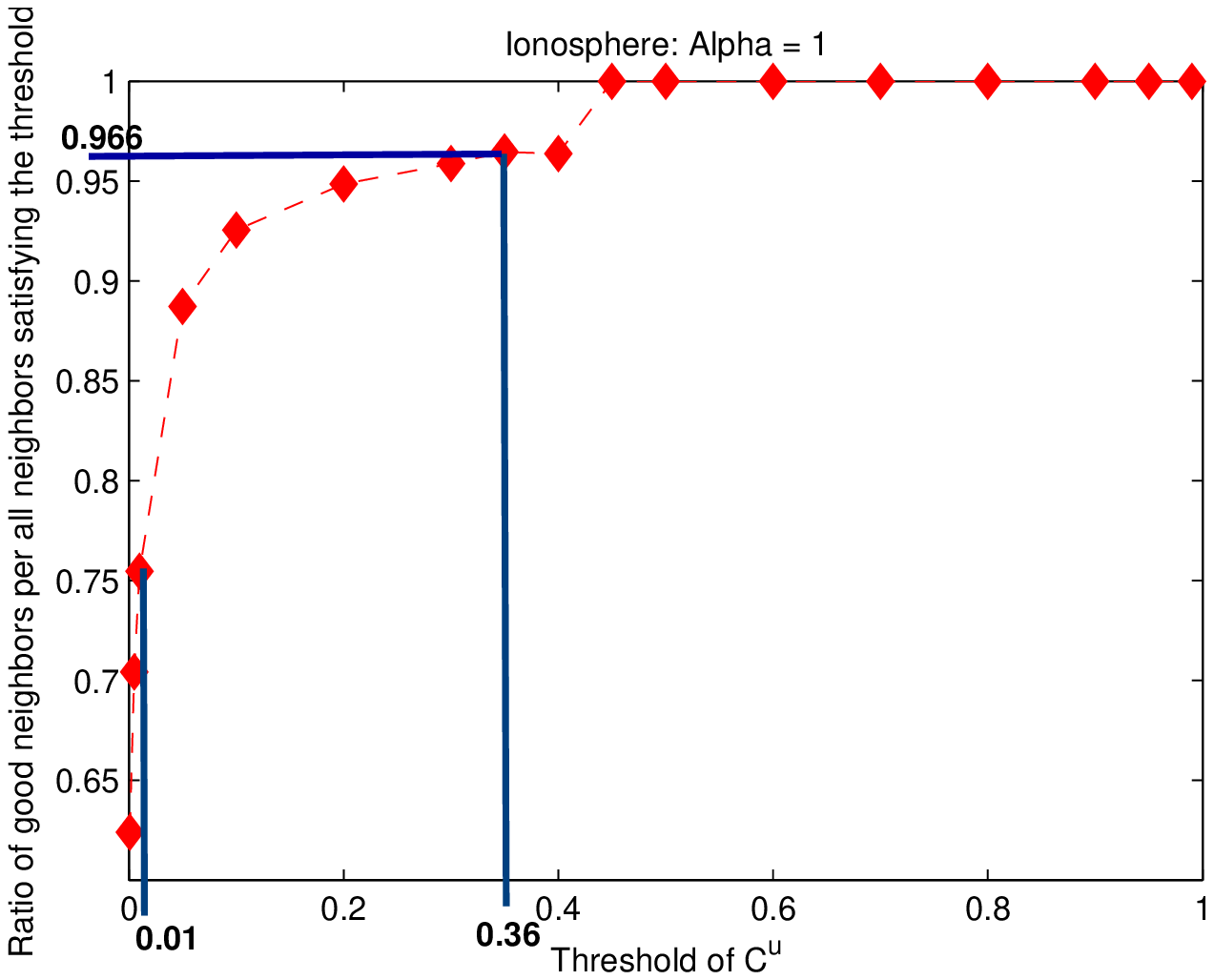}
\setlength{\epsfxsize}{3.3in} \epsfbox{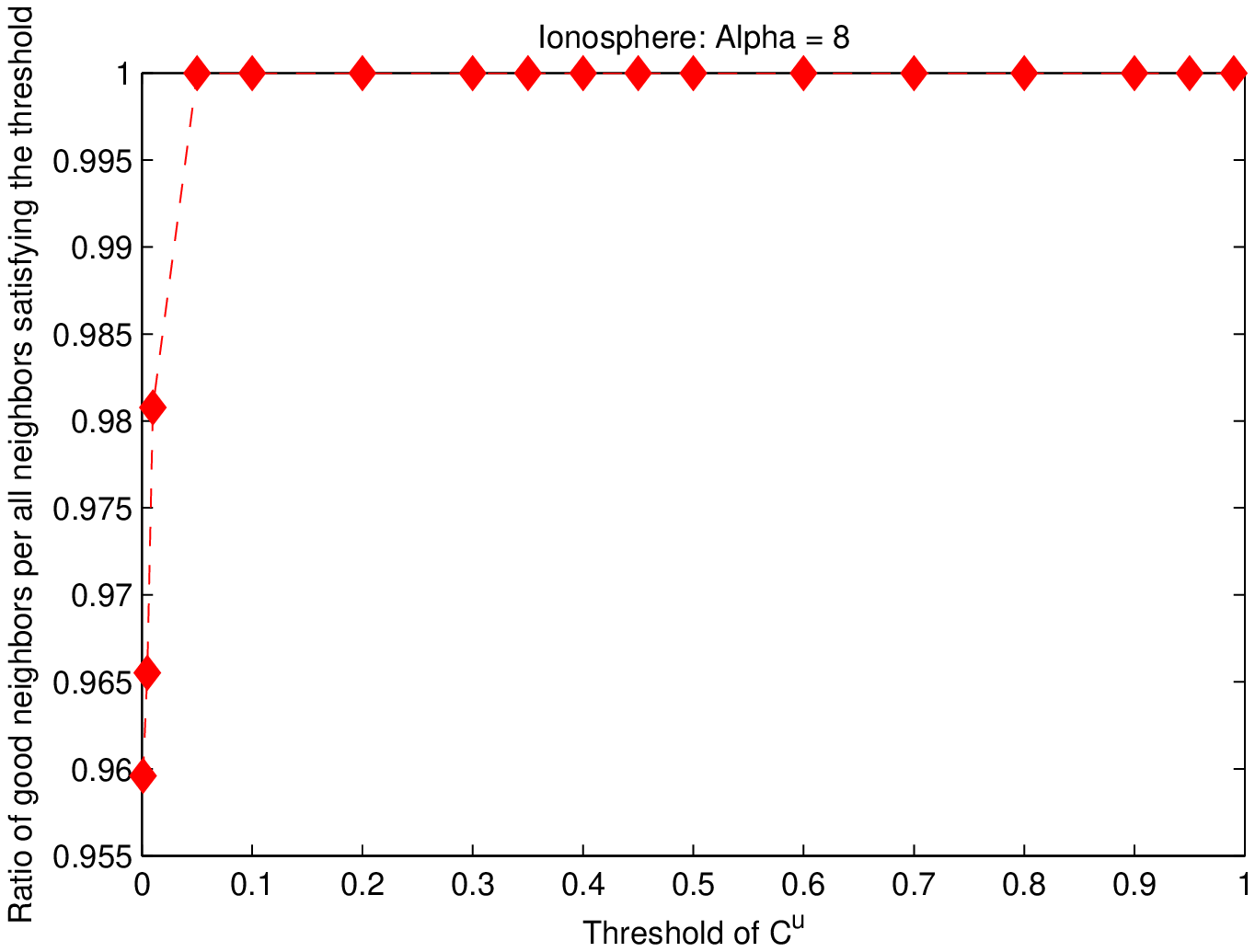} }
\end{center}
\vskip -0.5in \caption{For each number $x$ in the the x-axis, its
corresponding value on the y-axis is the \emph{ratio between the
number of good nearby examples} (having $c^u_{ij} > x$ and belonging
to the same class) \emph{and the number of nearby examples} (having
$c^u_{ij} > x$). The ratios with respect to $C^{u^{\alpha}}$ are
demonstrated where (Left) $\alpha = 1$ (the standard LPP), and where
(Right) $\alpha = 8$ (LPP*).} \label{fig_ionos_hadam} \vskip -0.0in
\end{figure*}

It can be observed that LPP* is much better than PCA on this
dataset, and therefore, unlike SELF, SS-LFDA much improves LFDA. In
fact, the main reason that SS-LFDA, SS-DNE and LPP* have good
classification performances are because of the Hadamard power
operator. This is explained in Figures~\ref{fig_simiarlity_ionos1},
\ref{fig_simiarlity_ionos2} and \ref{fig_ionos_hadam}. From
Figures~\ref{fig_simiarlity_ionos1} and \ref{fig_simiarlity_ionos2},
defining ``nearby examples'' be a pair of examples with a link
(having $c_{ij}^u \ge 0.36$), we see that\emph{ almost every link
connects nearby examples of the same class} (i.e. connects good
nearby examples). This indicates that our unlabel cost matrix $C^u$
is quite accurate as bad nearby examples rarely have links. In fact,
the \emph{ratio of good nearby examples per total nearby examples}
(shortly, the good-nearby-examples ratio) is 394/408 $\approx$
0.966. Nevertheless, if we re-define ``nearby examples'' be a pairs
of examples having, e.g., $c_{ij}^u \ge 0.01$, the same ratio then
reduces to $0.75$ as shown in Figure~\ref{fig_ionos_hadam} (Left).
This indicates that many pairs of examples having small values of
$c^u_{ij}$ are of different classes (i.e. bad nearby examples).

Since an algorithm derived from our framework minimizes the
\emph{cost-weighted average} distances of every pair of examples
(see Eq.~\eqref{eq_linear2} and its derivation), it is beneficial to
further increases the cost of a pair having large $c_{ij}^u$ (since
it usually corresponds to a pair of the same class) and decreases
the cost of of a pair having small $c_{ij}^u$. From
Eq.~\eqref{eq_hadamard_pow}, it can be easily seen that the effect
of the Hadamard power operator is exactly what we need. The
good-nearby-examples ratios after applying the Hadamard power
operator with $\alpha = 8$ are illustrated in
Figure~\ref{fig_ionos_hadam} (Right). Notice that, after applying
the operator, even pairs with small values of $c^u_{ij}$ are usually
of the same class.

%Note that as explained by an example shown in Figure~\ref{fig_toy3},
%the performance of SS-DNE and SS-LFDA can exceed their supervised
%and unsupervised counterparts.

%Note that ``0.36'' is just a random chosen number. It is chosen with
%the purpose that the number of links is not so many so that they are
%visualizable to human. Similar ratios of ``good nearby examples per
%total nearby examples'' according to a value $c^u_{ij} > x$ is shown
%in Figure~\ref{fig_ionos_hadam} (Left).

%To summarize, after applying the operator, high-score costs

\subsubsection{Balance}
\textsc{Balance} is an artificial dataset which was generated to
model psychological experimental results. Each example is classified
as having the balance scale tip to the right, tip to the left, or be
balanced. The 4 attributes containing integer values from 1 to 5 are
\textsc{left\_weight}, \textsc{left\_distance},
\textsc{right\_weight}, and \textsc{right\_distance}. The correct
way to find the class is the greater of (\textsc{left\_distance}
$\times$ \textsc{left\_weight}) and (\textsc{right\_distance}
$\times$ \textsc{right\_weight}). If they are equal, it is balanced.
Therefore, there are $5^4 = 625$ total examples and 3 classes in
this dataset. Moreover, the correct decision surface is
1-dimensional manifold lying in the feature space corresponding to
the $\inner{\cdot}{\cdot}^2$ kernel so that we set the target
dimensionality $d = 1$.

This dataset illustrates another flaw of using PCA in a
classification task. After centering, the covariance matrix of the
625 examples is just a multiple of $I$, the identity matrix.
Therefore, any direction is a principal component with largest
variance, and PCA is just return a random direction! Hence, we
cannot expect much about the classification performance of PCA in
this dataset. Thus, PCA cannot help SELF improves much the
performance on LFDA, and sometimes SELF degrades the performance of
LFDA due to overfitting. In contrast, SS-LFDA often improves the
performance of LFDA. Also, SS-DNE is able to improve the
classification performance of DNE and LPP* in all settings.

%NOTE: pca and balance. lpp and the first component. kdne and klfda
%will not get a perfect score.

\subsubsection{BCI}
This dataset originates from the development of a Brain-Computer
Interface where a single person performed 400 trials in each of
which he imagined movements with either the left hand (the $1^{st}$
class) or the right hand (the $2^{nd}$ class). In each trial,
electroencephalography (EEG) was recorded from 39 electrodes. An
autoregressive model of order 3 was fitted to each of the resulting
39 time series. The trial was represented by the total of 117 = 39*3
fitted parameters. The target dimensionality is set to the number of
classes, $d = c = 2$. Similar to the previous datasets, SS-LFDA and
SS-DNE are usually able to outperform their opponents. Again, PCA is
not appropriate for this real-world dataset, and hence SELF is
inferior to SS-LFDA.

\subsubsection{USPS}
This benchmark is derived from the famous USPS dataset of
handwritten digit recognition. For each digit, 150 images are
randomly drawn. The digits `2' and `5' are assigned to the first
class, and all others form the second class. To prevent a user to
employ a domain knowledge of the data, each example is rescaled,
noise added, dimension masked and pixel shuffled \cite[Chapter
21]{Semi:BOOK06}. Although there are only 2 classes in this dataset,
the original data presumably form 10 clusters, one for each digit.
Therefore, the target dimension $d$ is set to 10.

Often, SS-LFDA and SS-DNE outperform their opponents. Nevertheless,
note that SS-LFDA and SS-DNE do not improve much on LFDA and DNE
when $\ell = 100$ because 100 labeled examples are quite enough to
discriminating the data and therefore unlabeled examples offer
relatively small information to semi-supervised learners.

\subsubsection{M-Eyale}
This face recognition dataset is derived from \emph{extended Yale B}
\cite{Kriegman:PAMI01}. There are 28 human subjects under 9 poses
and 64 illumination conditions. In our \textsc{M-Eyale} (Modified
Extended Yale B), we randomly chose ten subjects, 32 images per each
subject, from the original dataset and down-sampling each example to
be of size 21$\times$24 pixels.

%\begin{figure}[b]
%\begin{center}
%\vskip -0.0in \setlength{\epsfxsize}{3.75in}
%\centerline{\epsfbox{eyale_crop2.eps}} \vskip -0.05in
%\caption{Extended Yale B Face dataset. 21 examples images from
%various illumination conditions.} \label{fig_eyale} \vskip -0.3in
%\end{center}
%\end{figure}

\textsc{M-Eyale} consists of 5 classes where each class consists of
images of two randomly-chosen subjects. Hence, there should be two
separated clusters for each class, and we should be able to see the
advantage of algorithms employing the conditions (1*) and (2*)
explained in Section~\ref{sect_cost}. In this dataset, the number of
labeled examples of each class is fixed to $\frac{\ell}{c}$ so that
examples of all classes are observed. Since this dataset should
consist of ten clusters, the target dimensionality is set to $d =
10$.

It is clear that LPP* performs much better than PCA in this dataset.
Recall that PCA captures maximum-variance directions; nevertheless,
in this face recognition task, maximum-variance directions are not
discriminant directions but directions of lighting and posing
\cite{Belhumeur:PAMI97}. Therefore, PCA captures totally wrong
directions, and hence PCA degrades the performance of SELF from
LFDA. In contrast, LPP* much better captures local structures in the
dataset and discover much better subspaces. Thus, by cooperating
LPP* with LFDA and DNE, SS-LFDA and SS-DNE are able to obtain very
good performances.

\section{Conclusion}
We have presented a unified semi-supervised learning framework for
linear and non-linear dimensionality reduction algorithms.
Advantages of our framework are that it generalizes existing various
supervised, unsupervised and semi-supervised learning frameworks
employing spectral methods. Empirical evidences showing satisfiable
performance of algorithms derived from our framework have been
reported on standard datasets.\\

\subsection*{Acknowledgements.} This work is supported by Thailand
Research Fund.

\bibliographystyle{unsrt}% bib style
\bibliography{SSSL_arxiv}% your bib database

\end{document}